%% file: main.tex
\documentclass[11pt, a4paper]{article}
\usepackage{fullpage}
\usepackage{amsfonts}
\usepackage{amssymb}
\usepackage{amsmath}
\usepackage{amsthm}
\usepackage{graphicx}
\usepackage{bm}
\usepackage[makeroom]{cancel}
\usepackage{enumitem}
\usepackage{url}
\usepackage[margin=.9in]{geometry}
\usepackage{amsopn}
\usepackage{mathtools}
\usepackage{hyperref}
\usepackage{doi}
\usepackage{cite}
\usepackage{overpic}
\usepackage[symbol]{footmisc}

\DeclareMathOperator*{\argmin}{arg\,min}

\usepackage{listings}
\usepackage{xcolor}
\definecolor{codegreen}{rgb}{0,0.6,0}
\definecolor{codegray}{rgb}{0.5,0.5,0.5}
\definecolor{codepurple}{rgb}{0.58,0,0.82}
\definecolor{backcolour}{rgb}{0.95,0.95,0.92}
\lstdefinestyle{mystyle}{
    backgroundcolor=\color{backcolour},   
    commentstyle=\color{codegreen},
    keywordstyle=\color{magenta},
    numberstyle=\tiny\color{codegray},
    stringstyle=\color{codepurple},
    basicstyle=\ttfamily\normalsize,
    breakatwhitespace=false,         
    breaklines=true,                 
    captionpos=b,                    
    keepspaces=true,                 
    numbers=left,                    
    numbersep=3pt,                  
    showspaces=false,                
    showstringspaces=false,
    showtabs=false,                  
    tabsize=4,
}
\renewcommand{\vec}[1]{\mathbf{#1}}

\newcommand{\norma}[1]{\left|\left|{#1}\right|\right|}

\newcommand{\scalarprod}[2]{\left({#1},\,{#2}\right)}

\newcommand{\dpart}[2]{\frac{\partial {#1}}{\partial {#2}}}

\newtheorem{definition}{Definition}

\begin{document}
\begin{center}
    \Large \bf \texttt{pyforce-1.0.0}: Python Framework for data-driven model Order Reduction of multi-physiCs problEms
\end{center}
\begin{center}
    Stefano Riva$^1$,
    Yantao Luo$^1$,
    Carolina Introini$^{1}$,
    Antonio Cammi$^{2,1}$
\end{center}
\begin{center}
    \scriptsize{
    ${}^1$ Department of Energy, Nuclear Engineering Division, Politecnico di Milano, Milan, Italy \\  
    ${}^2$ Department of Mechanical and Nuclear Engineering and Emirates Nuclear Technology Center, Khalifa University, Abu Dhabi (127788), United Arab Emirates}
\end{center}

\begin{abstract}
\texttt{pyforce} is a Python package implementing Data-Driven Reduced Order Modelling techniques for applications to multi-physics problems, mainly set in the Nuclear Engineering world. The package is part of the ROSE (Reduced Order modelling with data-driven techniques for multi-phySics problEms): mathematical algorithms aimed at reducing the complexity of multi-physics models (for nuclear reactors applications), at searching for optimal sensor positions and at integrating real measures to improve the knowledge on the physical systems.

With respect to the previous original implementation based on \texttt{dolfinx} package (v0.6.0), version 1.0.0 of \texttt{pyforce} has been completely re-written using pyvista as backend for mesh importing, computing integrals, and visualisation of results; in addition, functions are stored as numpy arrays, improving the ease of use of the package. This choice allows to use \texttt{pyforce} with any software solver able to export results in VTK format.
\end{abstract}

\noindent 
\textbf{Github Repository: } \url{https://github.com/ERMETE-Lab/ROSE-pyforce}\\
\textbf{Documentation: } \url{https://ermete-lab.github.io/ROSE-pyforce/intro.html}

\section{Introduction}
Data-Driven Reduced Order Modelling (DDROM) techniques have gained significant attention in recent years due to their ability to efficiently reduce high-dimensional spatial-temporal data and learn accurate surrogate models for complex dynamical systems \cite{quarteroni_reduced_2015, lassila_model_2014}. These methods leverage data to identify low-dimensional representations of the underlying dynamics, enabling faster simulations and predictions while preserving essential features of the original system. These methods become particularly relevant for multi-query and real-time applications, where traditional high-fidelity models, i.e. those based on the discretization of Partial Differential Equations (PDEs), can be computationally prohibitive. In fact, in optimization, uncertainty quantification, or monitoring tasks is important to have fast yet accurate models that can provide reliable predictions in a reasonably short time.
In this context, \texttt{pyforce} \cite{pyforce_JOSS, RIVA2024_AMM, CAMMI2024_NED} is a Python package that implements DDROM techniques for multi-physics problems, with a focus on applications in Nuclear Engineering. The package is part of the ROSE (Reduced Order modelling with data-driven techniques for multi-phySics problEms) project, which aims to develop mathematical algorithms for reducing the complexity of multi-physics models, optimizing sensor placement, and integrating real measurements to enhance the understanding of physical systems.

From the implementation point of view, the original version of \texttt{pyforce} (v0.1.3) was based on the \texttt{dolfinx} package \cite{BarattaEtal2023, ScroggsEtal2022, BasixJoss, AlnaesEtal2014}, which provided a powerful framework for finite element analysis and mesh handling. However, to improve the usability and flexibility of the package, especially in handling data coming from other codes, version 1.0.0 of \texttt{pyforce} has been completely re-written using \texttt{pyvista} \cite{sullivan2019pyvista} as the backend for mesh importing, computing integrals, and visualisation of results. This change allows users to work with any software solver capable of exporting results in VTK format, significantly broadening the applicability of \texttt{pyforce}. As an example, the \texttt{dolfinx} dependency required meshes to be created within the \texttt{dolfinx} framework and were constrained to have a single topological entity (e.g., only hexahedral or thetrahedral elements). With the new \texttt{pyvista} backend, users can import meshes created with various software tools, as soon as they are exported in VTK format, and can work with hybrid meshes containing different types of elements.

\section{Mathematical Background} \label{sec:background}
Let $\Omega\subset \mathbb{R}^d$, with $d=\{1,2,3\}$, be the spatial domain onto which the functional space $L^2(\Omega)$ is defined. This is a Hilbert Space endowed with an inner product and the induced norm:
\begin{equation}
	\scalarprod{u_1}{u_2}_{L^2(\Omega)} = \int_\Omega u_1\cdot u_2\,d\Omega \qquad\qquad \norma{u}_{L^2(\Omega)} = \sqrt{ \scalarprod{u}{u}_{L^2(\Omega)}}
\end{equation}
for any $ u_1, u_2\in L^2(\Omega)$. Let $\mathcal{U}\subset L^2(\Omega)$ be a Hilbert Space endowed with a proper inner product and induced norm, usually referred to as the \textit{solution manifold}. In this context, any problem can be studied and modelled by a parametrized Partial Differential Equation (PDE) in space and time, similar to the single-physics problems presented in the previous chapter, such as the following one:
\begin{equation}
	\dpart{u}{t} + \mathcal{G}(u; \vec{x}, t, \boldsymbol{\mu})=f(\vec{x}; t, \boldsymbol{\mu})\qquad\qquad
    \vec{x}\in\Omega,\; t\in\mathcal{T}
	\label{eqn-chap2: PDE}
\end{equation}
where $\mathcal{G}$ is a differential (or integro-differential) operator, $f$ is the forcing term, $\mathcal{T}\in\mathbb{R}^+_0$ is the time interval and $\boldsymbol{\mu}$ represents the parameters sampled in the domain $\mathcal{D}\subseteq\mathbb{R}^p$ ($p\geq 1$). This differential problem will be sometimes referred to as the \textit{best-knowledge} problem or \textit{background model}, whose solution lives in $\mathcal{U}$.

The solution manifold $\mathcal{U}$ is usually infinite-dimensional and unknowable; typically, the available information consists of a collection of solutions for some values of the parameters\footnote{Time plays a special role among the "parameters" since typically time-dependent derivatives are present within the starting equations, nevertheless is usually accounted among the parameters $\boldsymbol{\mu}$. Through the thesis, if $time$ is omitted from the notation it means that it is included in the parameter vector $\boldsymbol{\mu}$.} $(t, \boldsymbol{\mu})$, called \textit{snapshots} \cite{quarteroni_reduced_2015}. The aim of linear dimensionality reduction approaches, onto which ROM methods are built, consists of constructing an optimal finite-dimensional representation of $\mathcal{U}$, called \textbf{reduced space} or \textbf{latent space} \cite{brunton_data-driven_2022} and indicated with $X_N$, from the information contained in the snapshots. 

Adopting a geometrical analogy, the solution manifold $\mathcal{U}$ can be visualised as a non-planar surface, with the snapshots as some known points on it; reduction techniques then allows to find the optimal plane that best fits the different available points. The reduced space is spanned by the basis functions $\{\varphi_i(\vec{x})\}_{i=1}^N$, which describe the dominant spatial physics of the problem under study. In particular, any solution $u(\vec{x}; \boldsymbol{\mu})$, belonging to the solution manifold $\mathcal{U}$ can be expressed as a linear combination of the basis functions \cite{quarteroni_reduced_2015, rozza_model_2020}:
\begin{equation}
    u(\vec{x}; \boldsymbol{\mu}) \simeq \sum_{i=1}^N \alpha_i(\boldsymbol{\mu}) \cdot \varphi_i(\vec{x})
    \label{eqn: chap02-rom-linear-exp}
\end{equation}
in which the parametric dependence (including the temporal dynamics) is embedded in the coefficients $\{\alpha_i(\boldsymbol{\mu})\}_{i=1}^N$, sometimes called \textbf{reduced coefficients} or \textbf{latent dynamics}.

\subsection{Application to Data Assimilation}

The Data Assimilation (DA) paradigm {\cite{carrassi_data_2018} is a general mathematical framework providing the theoretical background behind the integration of background models and real data and, thus, for the state estimation problem:

\begin{definition}{\textbf{(State Estimation)}}{}
Given a best-knowledge multi-physics model $\mathcal{B}$ for $u^{\text{bk}}$ and $M$ local observations $\vec{y}\in\mathbb{R}^M$, develop an algorithm $\mathcal{A}$ that takes the measurements and returns an estimate $u^\star=\mathcal{A}(\vec{y}, \mathcal{B})$ of the true state $u^{\text{true}}$, exploiting the background model $\mathcal{B}$. The computational time to retrieve the estimate should be independent of the high-fidelity solver used to discretize the best-knowledge problem.
\end{definition}

A general variational formulation of it, also known as the DA statement, can be given: the aim of any DA technique consists in obtaining the optimal state estimation $u^\star$ that minimizes the distance between the background model $u^{bk}$ and the real observations $\vec{y}^{obs}$, i.e. 
\begin{equation}
	u^\star = \argmin\limits_{u\in\mathcal{U}}\xi \norma{u-u^{bk}}_{L^2}^2+\frac{1}{M}\sum_{m=1}^M\left(v_m(u)-y_m^{obs}\right)^2,
	\label{eqn: DA-variational}
\end{equation}
where {$\xi\in[0,+\infty)$} is called the regularising parameter, which weights the relative importance of the background mathematical model compared to the experimental data: in particular, its optimal value depends on the overall model uncertainty\footnote{This expression is used to indicate simplifying assumptions made while deriving the model, uncertainty on physical parameters because they cannot be accurately measured or the numerical accuracy.} and the noise level. The optimisation problem \eqref{eqn: DA-variational} looks for an estimation close enough to the mathematical model and enriched with the novel information from the data.

Due to the computational costs associated with the solution of this optimisation problem, especially due to the need for multiple evaluations of the background model. This is one of the main reasons why the optimisation problem is not feasible for online applications, for which the use of surrogate modelling methods becomes necessary \cite{cheng_torchda_2025}, either with linear or non-linear dimensionality reduction methods \cite{rozza_model_2020} or Neural Networks \cite{goodfellow_deep_2016}. For several applications, the former is a more reliable solution due to the higher interpretability of the basis functions, the fewer requirements for the amount of training data and less training computational costs. These methods allow for representing the system into a different coordinate system, called \textbf{latent/reduced space}, in which the DA problem can be tackled with intrinsically lower computational costs. This framework will be called, throughout the thesis, Data-Driven Reduced Order Modelling (DDROM) \cite{pyforce_JOSS, RIVA2024_AMM, CAMMI2024_NED}.

\begin{figure}[tp]
    \centering
    \includegraphics[width=1\linewidth]{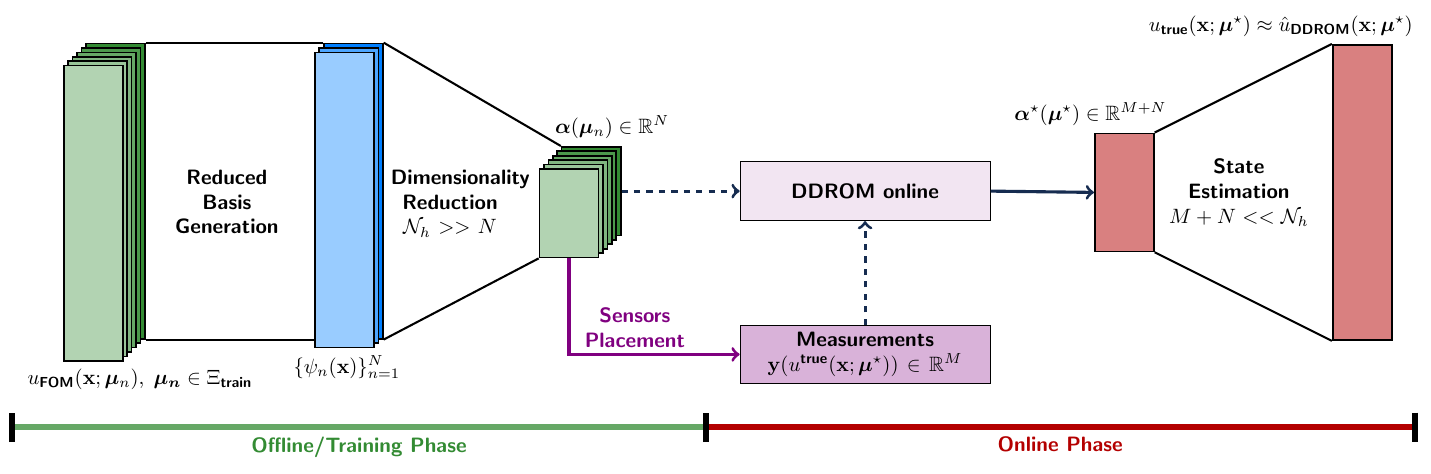}
    \caption{Scheme of Data-Driven Reduced Order Modelling for multi-physics problems implemented in \texttt{pyforce} \cite{RIVA2024_AMM}.}
    \label{fig: tie-fighter}
\end{figure}

This general framework includes different methods operating under the same paradigms and logic: in particular, the combination of models with real measures can increase the background knowledge provided by the PDEs with the information embedded inside the measures, representing the local evaluations of the true state. Focusing more specifically on Figure \ref{fig: tie-fighter}, two main phases are occurring, the offline (or training) and online (or testing) phases. The former includes the data generation, i.e. the retrieval of multiple solutions of the parametric PDEs and the generation of a proper set of basis functions allowing for a change in the coordinate system and hence a dimensionality reduction from $\mathcal{N}_h$ to $N$ degrees of freedom. This encoding procedure separates the spatial behaviour (within the basis functions) from the time/parametric behaviour (within the reduced coefficients); additionally, reasoning in the latent space helps in reducing the computational complexity of sensor placement methods. On the other hand, the online phase is characterised by quick evaluations of the state, with reasonably low computational costs: in particular, when a new set of measures becomes available, the surrogate model can be solved, ensuring the generation of an updated set of coordinate combining the background model and the information from the new local measures; this new set can be used to decode back to the high-dimensional space, providing a state estimation for monitoring, control or design purposes \cite{RIVA2024_AMM}.

\section{\texttt{pyforce} Structure and Usage}
 \label{sec:code}

At the moment, the following techniques have been implemented:

\begin{itemize}
    \item \textit{Singular Value Decomposition} (randomised, hierchical and incremental), with Projection and Interpolation for the Online Phase \\$\Rightarrow$ \texttt{pyforce.offline.pod, pyforce.online.pod}
    \item \textit{Proper Orthogonal Decomposition} with Projection and Interpolation for the Online Phase \\$\Rightarrow$ \texttt{pyforce.offline.pod, pyforce.online.pod}
    \item \textit{Empirical Interpolation Method}, either regularised with Tikhonov or not \\$\Rightarrow$ \texttt{pyforce.offline.eim, pyforce.online.eim}
    \item \textit{Generalised Empirical Interpolation Method}, either regularised with Tikhonov or not \\$\Rightarrow$ \texttt{pyforce.offline.geim, pyforce.online.geim}
    \item \textit{Parameterised-Background Data-Weak} formulation for Data Assimilation \\$\Rightarrow$ \texttt{pyforce.online.pbdw}
    \item \textit{SGreedy} algorithm for optimal sensor positioning \\$\Rightarrow$ \texttt{pyforce.offline.sgreedy}
    \item an \textit{Indirect Reconstruction} algorithm to reconstruct un-observable fields from observable ones \\$\Rightarrow$ \texttt{pyforce.online.indirect\_reconstruction}
\end{itemize}

This package is aimed to be a valuable tool for other researchers, engineers, and data scientists working in various fields, not only restricted in the Nuclear Engineering world.

\subsection{\texttt{tools} subpackage}

The \texttt{tools} subpackage includes utility functions for data manipulation, storage and import and other useful functions for the users, such as the train--test split function for the snapshots, scaling functions for the snapshots and the parameters, computing integrals on the grid, and so on. The most import class is \texttt{FunctionsList}, which is a custom list-like class to store the snapshots, with some useful methods for data manipulation and storage. Every field (either scalar, vector or tensor) is stored as a flattened numpy array, with the same ordering as the grid points. Furthermore, one of the main novelties of this version lies in the brand new classes to import cases from OpenFOAM \cite{OPENFOAM}, which is one of the most widely used CFD and Multi-Physics solvers in the Nuclear Engineering world. The classes are able to read both reconstructed and parallel-decomposed cases, single or multi-region, and to store the snapshots in the \texttt{FunctionsList} format, allowing for a seamless integration with the rest of the package. The following code snippet shows how to import an OpenFOAM case using the new classes:

\begin{lstlisting}[language=Python]
from pyforce.tools.write_read import ReadFromOF

path_to_case = 'path/to/your/case'

of = ReadFromOF(path_to_case)

# Read the case grid
mesh = of.mesh()

# Import temperature and velocity fields as snapshots
temperature_snaps, time_instants = of.import_field('T')
velocity_snaps, time_instants = of.import_field('U')

\end{lstlisting}

\subsection{\texttt{offline} subpackage}

The classes within the \texttt{offline} subpackage are responsible for the dimensionality reduction phase, which includes the generation of the spatial basis functions, and the search for optimal sensor positions. The dimensionality reduction techniques implemented in \texttt{pyforce} are based on linear methods: this choice is mainly motivated by the fact that linear methods are more interpretable, require less training data and have lower training computational costs compared to non-linear methods, such as Neural Networks. For what concerns the optimal sensor positioning, the implemented methods are based on greedy procedures. Figure \ref{fig: offline-classes} shows how the classes in the \texttt{offline} subpackage are structured: a set of training snapshots is given as input and are needed to generate the basis functions and to search for optimal sensor positions. The output of this phase consists of the basis functions, the optimal sensor positions and the projection matrix, which are needed for the online phase.

\begin{figure}[h!]
    \centering
    \includegraphics[width=1\linewidth]{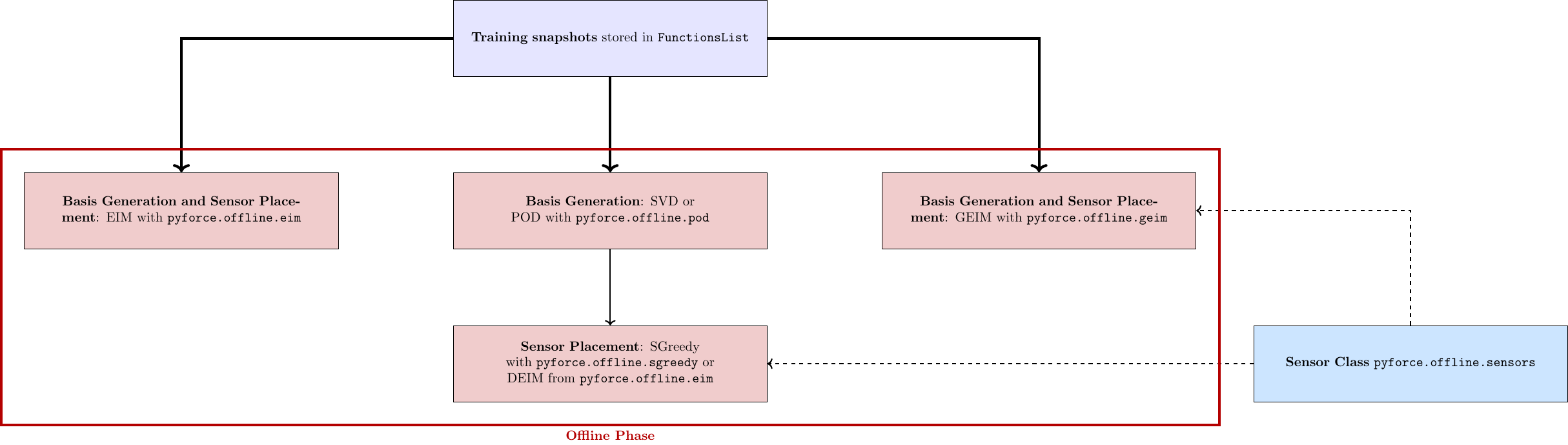}
    \caption{Scheme on how the classes in the \texttt{offline} subpackage are structured and how they interact with each other.}
    \label{fig: offline-classes}
\end{figure}

\subsection{\texttt{online} subpackage}

Given the basis functions and the sensors placed in the offline phase, the objective becomes the reconstruction of the state of the system given a set of local measurements of some characteristic fields or the characteristic unseen parameter of the state. This latter approach cointains different non-intrusive ROM methodologies, ranging from interpolation methods towards machine and deep learning approaches. Figure \ref{fig: online-classes-param} shows how the state can be estimated using the POD with Interpolation \cite{tezzele_integrated_2022}.

\begin{figure}[h!]
    \centering
    \includegraphics[width=1\linewidth]{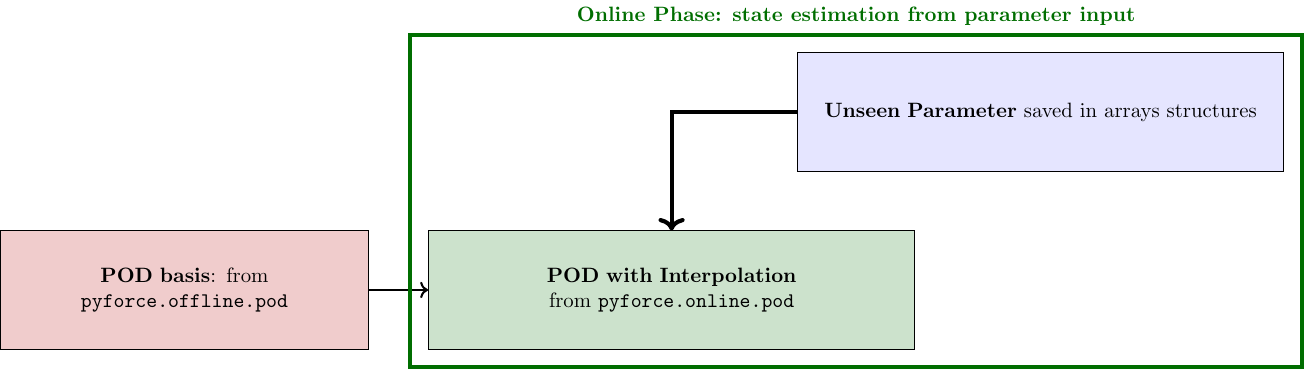}
    \caption{Scheme on how the classes in the \texttt{online} subpackage for non-intrusive reduced order modelling techniques are structured and how they interact with each other.}
    \label{fig: online-classes-param}
\end{figure}

Furthermore, as said, another possibility adopts sparse observations of some characteristic fields to reconstruct the state of the system, which is a more challenging problem falling into the category of inverse problems. As reported in \cite{CAMMI2024_NED,RIVA2024_AMM}, the quantities of interest in a nuclear reactor can be quite a few and not all of them can be directly observed; therefore, supposing to have two coupled fields to reconstruct $(\phi, \vec{u})$ and only local evaluations of $\phi$ are available, two different problems arise:

\begin{itemize}
    \item \textbf{Direct State Estimation}, Figure \ref{fig: online-classes-state-estimation}a:  from measurements of $\phi$, its spatial distribution has to be reconstructed (the information of $\vec{u}$ is not entering in this stage).
    \item \textbf{Indirect State Estimation}, Figure \ref{fig: online-classes-state-estimation}b: from measurements of $\phi$, the spatial distribution of $\vec{u}$ has to be reconstructed.
\end{itemize}

\begin{figure}[h!]
    \centering
    \begin{overpic}[width=1\linewidth]{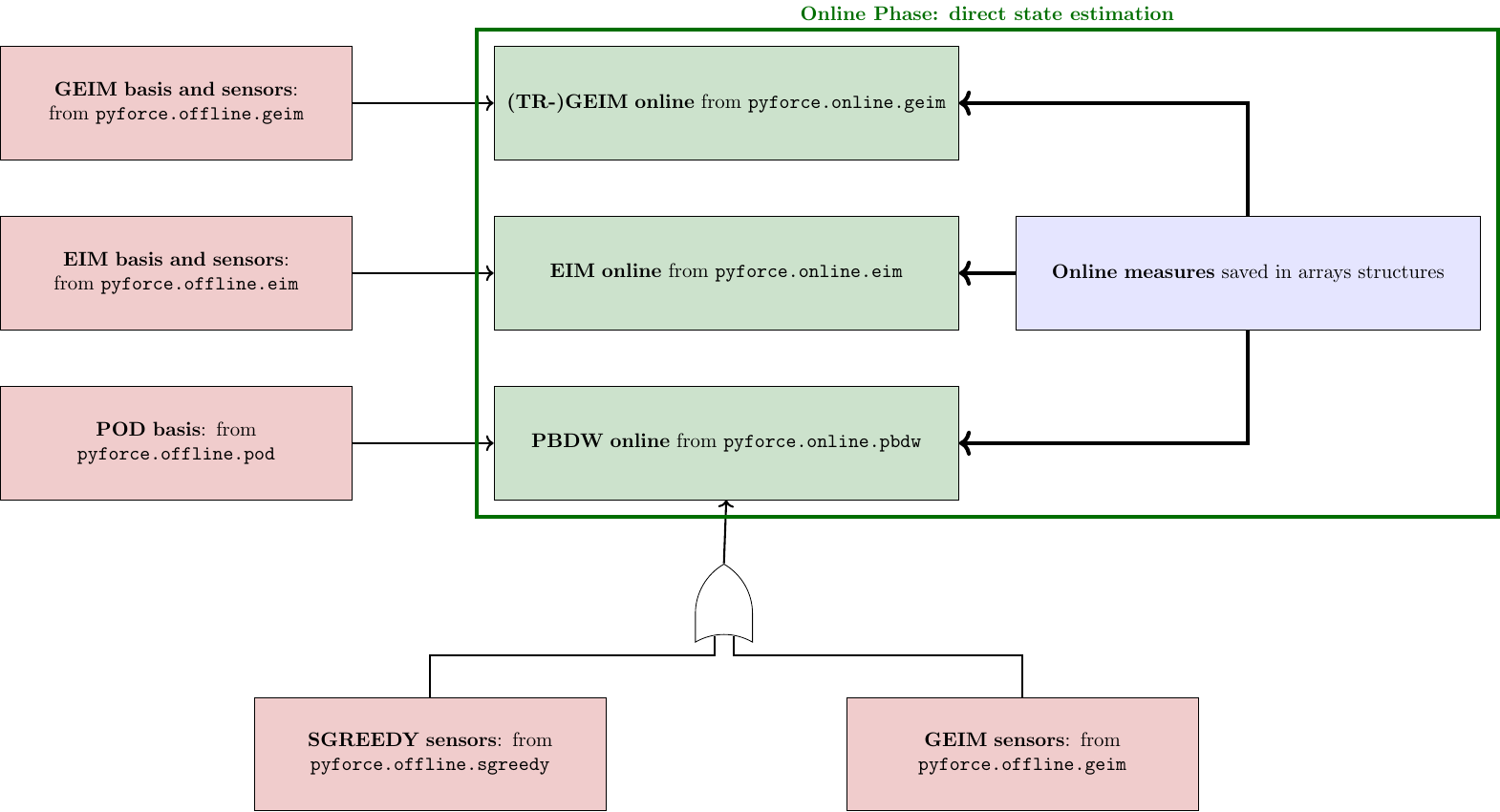}
    \put(0,53){(a)}
    \end{overpic}\vspace{0.5cm}
    \begin{overpic}[width=1\linewidth]{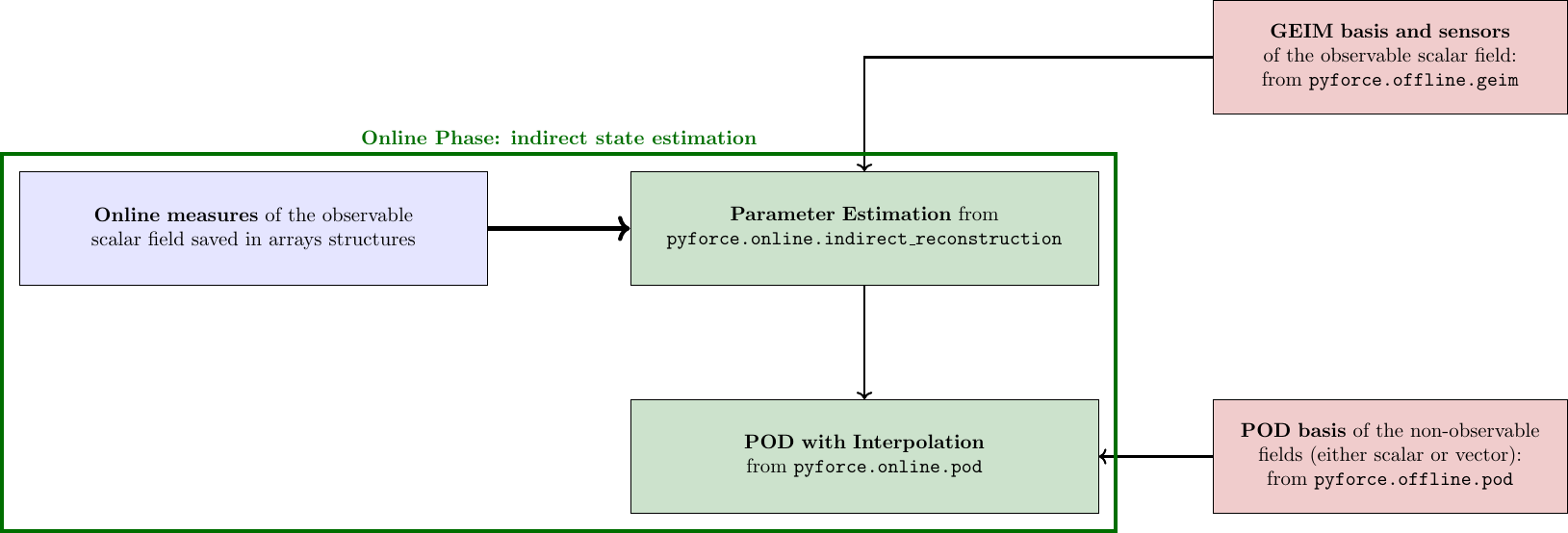}
    \put(0,26){(b)}
    \end{overpic}
    \caption{Scheme on how the classes in the \texttt{online} subpackage for direct (a)and indirect (b) reconstruction techniques are structured and how they interact with each other.}
    \label{fig: online-classes-state-estimation}
\end{figure}

\subsection{Example Usage}
Consider a grid defined on the $[0,1]^2$ domain and a toy function defined as
\begin{equation}
f(x,y;\mu) =
\sin(\pi \mu x)\,\cos(\pi \mu y)
+ \cos(\pi x)\,\sin\bigl((1-\mu)\pi y^2\bigr),
\end{equation}
where $\mu \in [-5,5]$ is a parameter. The following demo shows how to create the grid for this square domain, define the snapshots and assign them to the \texttt{FunctionsList} class, perform a train--test split, compute the Singular Value Decomposition (SVD), plot the singular values, and project a test snapshot onto the reduced space, i.e. $f\simeq\hat{f} =\sum_{i=1}^N \alpha_i \varphi_i$ with $\alpha_i = \scalarprod{f}{\varphi_i}$.

\paragraph{Grid Generation}

Square geometries can be created using the \texttt{pyvista.ImageData} class. The following code creates a grid with $50 \times 50$ elements on the $[0,1]^2$ domain.

\begin{lstlisting}[language=Python]
import pyvista as pv

nx = 50
ny = 50
nz = 1

grid = pv.ImageData(
    dimensions=(nx+1, ny+1, nz+1),
    spacing=(1/nx, 1/ny, 1e-4),
    origin=(0.0, 0.0, 0.0)
)
\end{lstlisting}

\paragraph{Snapshot Generation}

The snapshots are defined by evaluating the toy function on the grid for different values of the parameter $\mu$.

\begin{lstlisting}[language=Python]
from pyforce.tools.functions_list import FunctionsList
import numpy as np

# Extract the coordinates of the grid points
X, Y = grid.points[:, 0], grid.points[:, 1]

# Define the parameter values for which to generate snapshots
mu_values = np.linspace(-5, 5, 100)

# Define the toy function to generate snapshots
def harmonic_oscillator(X, Y, mu):
    term1 = np.sin(mu * np.pi * X) * np.cos(np.pi * Y)
    term2 = np.cos(np.pi * X) * np.sin((1 - mu) * np.pi * Y**2)
    return term1 + term2

# Create a FunctionsList to store the snapshots
snapshots = FunctionsList(dofs=len(X))
for mu in mu_values:
    snapshot = harmonic_oscillator(X, Y, mu)
    snapshots.append(snapshot)
\end{lstlisting}

\begin{figure}[h!]
\centering
\includegraphics[width=1\linewidth]{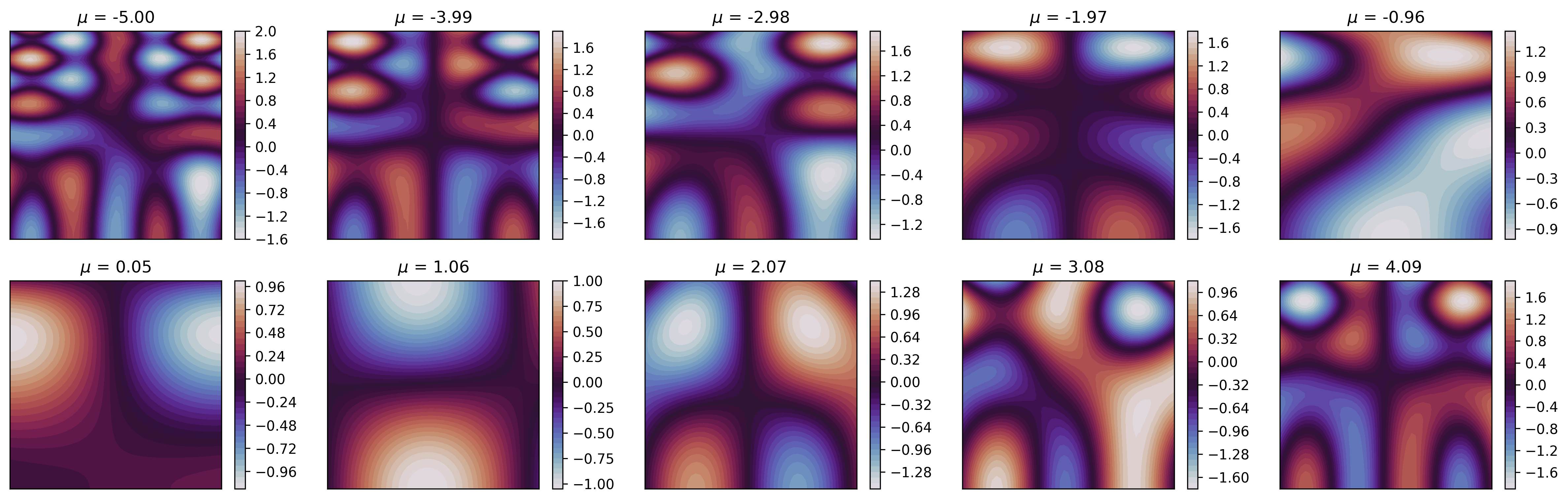}
\caption{Some of the snapshots generated from the toy function for different values of $\mu$.}
\end{figure}

\paragraph{Train--Test Split}

The dataset is split into training and testing sets as follows:

\begin{lstlisting}[language=Python]
from pyforce.tools.functions_list import train_test_split

train_mu, test_mu, train_snaps, test_snaps = train_test_split(
    mu_values, snapshots, test_size=0.2, random_state=42
)
\end{lstlisting}

\paragraph{Reduced-Order Model Construction}

The SVD is computed on the training snapshots, retaining $20$ modes.

\begin{lstlisting}[language=Python]
from pyforce.offline.pod import rSVD

svd = rSVD(grid, gdim=3)
svd.fit(train_snaps, rank=20)
eig_fig = svd.plot_sing_vals()
\end{lstlisting}

The class \texttt{rSVD} implements a method for plotting the singular values: in particular, their raw decay is plotted, together with the cumulative and residual energy content, which are useful to decide how many modes to retain for the reduced-order model construction (Figure \ref{fig: singular-values}).

\begin{figure}[h!]
\centering
\includegraphics[width=1\textwidth]{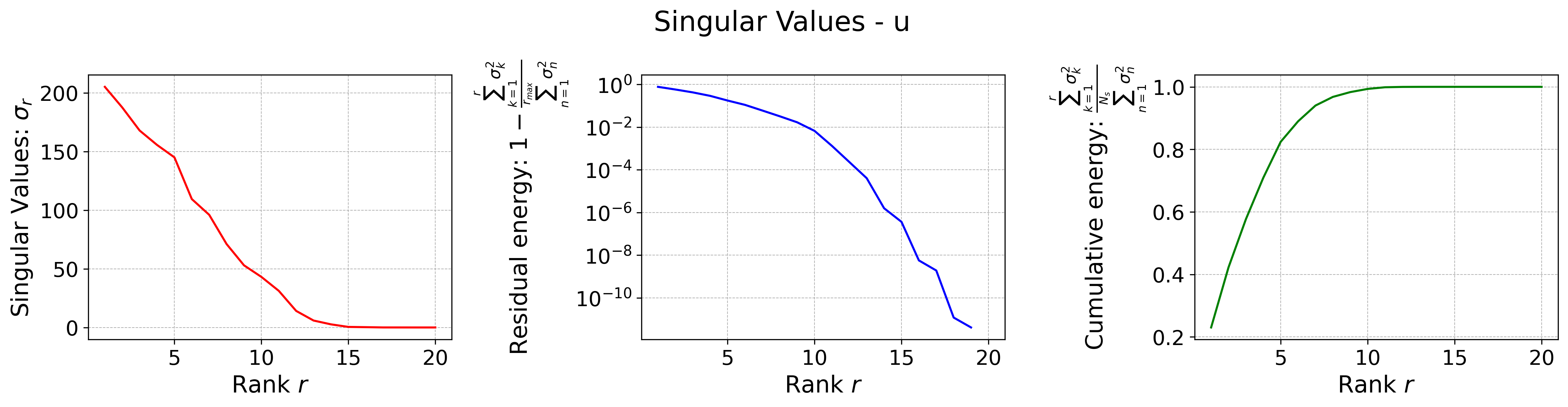}
\caption{Singular values of the snapshot matrix.}
\label{fig: singular-values}
\end{figure}

\paragraph{Projection and Reconstruction}

Finally, a test snapshot is projected onto the reduced space and reconstructed:

\begin{lstlisting}[language=Python]
test_index = 0  # Index of the test snapshot to project
test_snapshot = test_snaps[test_index]
reduced_coeffs = svd.project(test_snapshot)
reconstructed_snapshot = svd.reconstruct(reduced_coeffs)
\end{lstlisting}

\begin{figure}[h!]
\centering
\includegraphics[width=\textwidth]{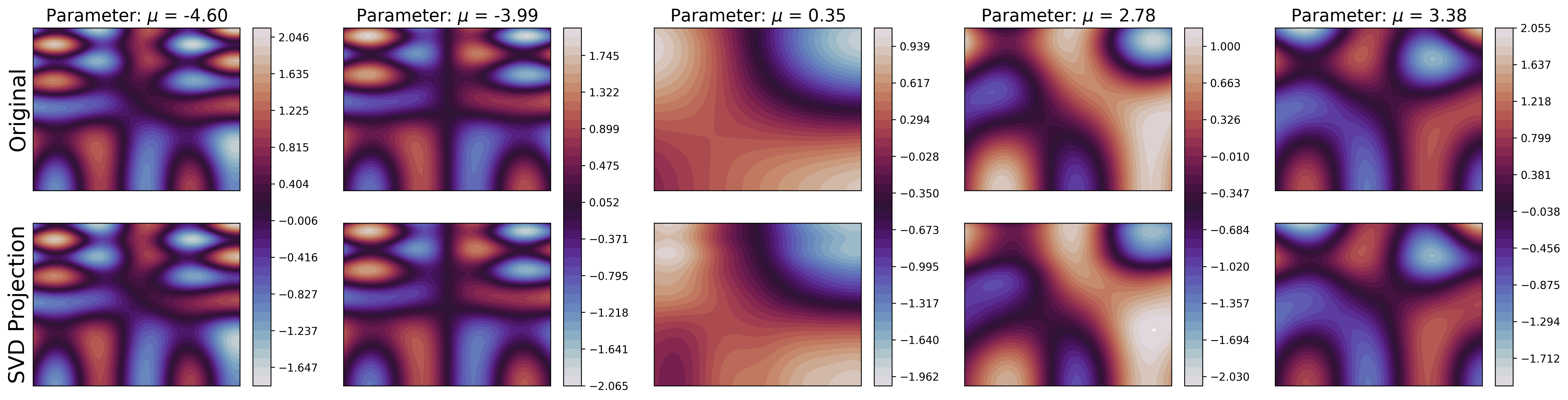}
\caption{Comparison between the original (first row) and reconstructed (second row) test snapshot.}
\end{figure}
\section{Conclusion}
This paper presented the novel features of version 1.0.0 of \texttt{pyforce}, a Python package for Data-Driven Reduced Order Modelling of multi-physics problems, with a focus on applications in Nuclear Engineering. The package has been completely re-written using \texttt{pyvista} as the backend for mesh importing, computing integrals, and visualisation of results, allowing for greater flexibility and ease of use. The package includes various techniques for dimensionality reduction and optimal sensor placement, as well as methods for state estimation from local measurements. An example usage was provided to demonstrate how to create a grid, generate snapshots, perform a train--test split, compute the SVD, and project a test snapshot onto the reduced space. The appendix includes more detailed examples of the usage of the package for different techniques and applications (also available on the official documentation of the package).


\bibliographystyle{unsrtsiam}
\bibliography{refs}

\newpage

\input{appendix}

\end{document}

%% file: appendix.tex
\appendix
\section{Appendix: examples/tutorials}

\subsection{Introduction to SVD methods for Reduced Order Modelling}

This appendix provides a brief tutorial on the basic concepts of Reduced Order Modelling (ROM), utilizing Singular Value Decomposition (SVD). Crucially, this section highlights the features of the new version of the \texttt{pyforce} package, demonstrating its capabilities in importing data directly from OpenFOAM simulations, compressing fluid dynamics fields, and constructing surrogate models.

The workflow involves:
\begin{itemize}
    \item Importing snapshots from OpenFOAM simulations using the newly implemented \texttt{ReadFromOF} class.
    \item Compressing fluid dynamic fields (e.g., pressure and velocity) using randomised SVD and Proper Orthogonal Decomposition (POD) techniques.
    \item Creating surrogate models for reduced dynamics through interpolation and regression SVD.
\end{itemize}

The data utilized in the following examples originates from an OpenFOAM tutorial (\textit{pimpleFoam}, laminar conditions) known as \texttt{cylinder2D}.

\subsubsection{Reading OpenFOAM Cases}
A key feature of the new \texttt{pyforce} package is the \texttt{ReadFromOF} class, designed to easily import simulation data directly from OpenFOAM case folders.

\begin{lstlisting}[language=Python]
from pyforce.tools.write_read import ReadFromOF
import warnings
warnings.filterwarnings("ignore", category=RuntimeWarning)

io_data = ReadFromOF('Datasets/LaminarFlowCyl_OF2012/', 
                    skip_zero_time=True # optional
                    )
\end{lstlisting}

We can import target variables such as pressure ('p', scalar field) and velocity ('U', vector field). The \texttt{import\_field} method loads the data, returning the snapshots as a \texttt{FunctionsList} object along with a vector of time instances. While \texttt{pyvista} is the default loading library, alternatives like \texttt{fluidfoam} or \texttt{foamlib} can be used.

\begin{lstlisting}[language=Python]
import numpy as np

var_names = ['p', 'U']
dataset = dict()

for field in var_names:
    print('Importing '+field+f' using pyvista')
    dataset[field], times = io_data.import_field(field,
                                              import_mode='pyvista', verbose=False # optional
                                              )
\end{lstlisting}

To facilitate snapshot plotting, the underlying grid required by \texttt{pyforce} is extracted using the \texttt{mesh} method.

\begin{lstlisting}[language=Python]
grid = io_data.mesh()
\end{lstlisting}

Additional derived fields, like the vorticity field ($\boldsymbol{\omega} = \nabla \times \mathbf{u}$), can be calculated and appended.

\begin{lstlisting}[language=Python]
from pyforce.tools.backends import LoopProgress
from pyforce.tools.functions_list import FunctionsList

var_names.append('vorticity')
mesh = io_data.mesh

dataset['vorticity'] = FunctionsList(dofs = dataset['U'].fun_shape)
bar = LoopProgress(final=len(times), msg='Computing vorticity field')
for ii, t in enumerate(times):
    grid['U'] = dataset['U'][ii].reshape(-1, 3)
    dataset['vorticity'].append( 
                        grid.compute_derivative(scalars='U',
                                                vorticity=True)['vorticity'].flatten()                    
                    )
    bar.update(1)
\end{lstlisting}

\subsubsection{Reduced Basis Construction: Singular Value Decomposition (SVD)}
We construct reduced bases using SVD and Proper Orthogonal Decomposition (POD). SVD decomposes a snapshot matrix $\mathbb{S}$ into $\mathbb{U} \Sigma \mathbb{V}^T$, yielding spatial modes ($\mathbb{U}$), temporal/parametric modes ($\mathbb{V}$), and their importance ($\Sigma$).

The \texttt{pyforce} package implements a highly efficient randomized version of SVD (\texttt{rSVD}), ideal for large datasets.

\begin{lstlisting}[language=Python]
from pyforce.offline.pod import rSVD

rsvd = dict()

for field in var_names:
    rsvd[field] = rSVD(grid, 
                       varname=field # optional
                      )
\end{lstlisting}

The \texttt{fit} method computes the modes and singular values by supplying the training snapshots (as a \texttt{FunctionsList}) and the desired rank.

\begin{lstlisting}[language=Python]
max_rank = 100

for field in var_names:
    rsvd[field].fit(train_datasets[field], rank=max_rank,
                    verbose=True # optional
                   )
\end{lstlisting}

\subsubsection{Surrogate Modeling for Latent Space}
Surrogate models describe the evolution of the temporal coefficients (latent space). In this tutorial, a Gaussian Process Regressor (GPR) model is defined using \texttt{sklearn} and wrapped within the \texttt{SurrogateModelWrapper} class.

\begin{lstlisting}[language=Python]
from sklearn.gaussian_process import GaussianProcessRegressor
from sklearn.gaussian_process.kernels import RBF, ConstantKernel as C
from sklearn.preprocessing import MinMaxScaler, StandardScaler
from IPython.display import clear_output as clc

class SurrogateGPR(SurrogateModelWrapper):
    def __init__(self):
        pass
    def fit(self, train_times, train_coeffs, scale_data=True, alpha=1e-4, n_restarts_optimizer=5, **kwargs):

        train_times = train_times.reshape(-1, 1)

        if scale_data:
            self.input_scaler = MinMaxScaler()
            self.input_scaler.fit(train_times)
            self.output_scalers = [StandardScaler() for _ in range(train_coeffs.shape[0])]
            for i in range(train_coeffs.shape[0]):
                self.output_scalers[i].fit(train_coeffs[i].reshape(-1, 1))
        else:
            self.input_scaler = None
            self.output_scalers = None

        self._models = []
        for i in range(train_coeffs.shape[0]):
            kernel = C(1.0, (1e-3, 1e3)) * RBF(length_scale=1.0, length_scale_bounds=(1e-2, 1e2))
            model = GaussianProcessRegressor(kernel=kernel, n_restarts_optimizer=n_restarts_optimizer, alpha=alpha)

            if scale_data:
                model.fit(self.input_scaler.transform(train_times), self.output_scalers[i].transform(train_coeffs[i].reshape(-1, 1)), **kwargs)
            else:
                model.fit(train_times, train_coeffs[i])
            self._models.append(model)

    def predict(self, test_times):
        if self.input_scaler is not None:
            test_times = self.input_scaler.transform(test_times.reshape(-1, 1))

        predictions = np.array([model.predict(test_times.reshape(-1,1)).flatten() for model in self.models])

        if self.output_scalers is not None:
            predictions = np.array([self.output_scalers[i].inverse_transform(predictions[i].reshape(-1,1)).flatten() for i in range(predictions.shape[0])])

        return predictions
    
    def predict_std(self, test_times):
        if self.input_scaler is not None:
            test_times = self.input_scaler.transform(test_times.reshape(-1, 1))

        stds = np.array([model.predict(test_times.reshape(-1,1), return_std=True)[1].flatten() for model in self.models])

        if self.output_scalers is not None:
            stds = np.array([self.output_scalers[i].scale_ * stds[i] for i in range(stds.shape[0])])

        return stds
    
for field in var_names:
    surr_models['GPR'][field] = SurrogateGPR()
    surr_models['GPR'][field].fit(train_times[train_idx], pod_coeffs[field][:max_test_rank, train_idx], scale_data=True)
\end{lstlisting}

\subsubsection{Error Analysis}
Finally, the \texttt{compute\_errors} method of the \texttt{POD} class is used to evaluate the reconstruction error for the test set, providing absolute and relative errors alongside computational costs.

\begin{lstlisting}[language=Python]
test_res = dict()

for field in var_names:
    test_res[field] = dict()
    for _meth in surr_models.keys():
        test_res[field][_meth] = online_pod[field].compute_errors(test_datasets[field],
                                                                  coeff_model = surr_models[_meth][field], 
                                                                  input_vector = test_times,
                                                                  verbose=True # optional
                                                                 )
\end{lstlisting}

\subsection{(Generalised) Empirical Interpolation Method - (G)EIM}

This section presents the basic applications of the (Generalised) Empirical Interpolation Method ((G)EIM) for dimensionality reduction and sensor placement in fluid dynamics problems. The (G)EIM is a powerful technique that approximates complex functions using a reduced number of basis functions and interpolation points (sensors).

The workflow involves:
\begin{itemize}
    \item Building \texttt{FunctionsList} objects directly from spatio-temporal matrices.
    \item Utilizing the EIM and GEIM classes to perform greedy algorithms for dimensionality reduction and sensor placement.
    \item Reconstructing the full field from sparse measurements, including the capability to assess the impact of random noise.
\end{itemize}

The data utilized in the following examples originates from a steady-state natural convection simulation of a Buoyant Cavity performed in OpenFOAM.

\subsubsection{Data Import and Preprocessing}
While \texttt{pyforce} provides dedicated OpenFOAM readers, it is also highly compatible with standard data arrays. Snapshots previously exported to an \texttt{npz} file can be effortlessly loaded. The target variable in this example is the temperature field ($T$).

\begin{lstlisting}[language=Python]
import numpy as np
from pyforce.tools.functions_list import FunctionsList

# Load parametric data and snapshots
_data = np.load('Datasets/BuoyantCavity_OF2412/buoyant_cavity.npz', allow_pickle=True)
parameters = _data['parameters']
field = 'T'

# Build the FunctionsList object directly from the snapshot matrix
dataset = FunctionsList(snap_matrix = _data[field] - 300)
\end{lstlisting}

To rigorously evaluate the methods, the dataset is partitioned into training and testing subsets using the built-in \texttt{train\_test\_split} utility.

\begin{lstlisting}[language=Python]
from pyforce.tools.functions_list import train_test_split

train_params, test_params, train_dataset, test_dataset = train_test_split(
    parameters, dataset, test_size=0.2, random_state=42
)
\end{lstlisting}

\subsubsection{Offline Phase: Sensor Placement and Basis Generation}
The offline phase is computationally intensive but is executed only once. It employs a greedy algorithm to iteratively select the most informative basis functions and optimal sensor locations (interpolation points). 

The \texttt{OfflineEIM} class is initialized with the computational grid and the variable name. The \texttt{fit} method subsequently computes the magic functions and interpolation points up to a maximum rank (\texttt{Mmax}).

\begin{lstlisting}[language=Python]
from pyforce.offline.eim import EIM as OfflineEIM

# Initialize and fit the Offline EIM class
eim_offline = OfflineEIM(grid, varname=field)
max_abs_err, beta_coeffs = eim_offline.fit(train_dataset, Mmax = 50, verbose=True)
\end{lstlisting}

The quality and stability of the EIM interpolation procedure can be assessed by computing the Lebesgue constant, a feature directly available within the class.

\begin{lstlisting}[language=Python]
lebesgue_eim = eim_offline.compute_lebesgue_constant()
\end{lstlisting}

\subsubsection{Online Phase: Field Reconstruction and Noise Handling}
During the online phase, the system state is reconstructed using sparse measurements. The \texttt{pyforce} package streamlines this by allowing users to import the basis functions and sensors calculated during the offline phase, and subsequently computing the interpolation matrix $B$.

\begin{lstlisting}[language=Python]
# Pass the pre-computed basis functions and sensors to the online class
geim_online.set_basis(basis = geim_offline.magic_functions)
geim_online.set_magic_sensors(sensors = geim_offline.magic_sensors.library)

# Compute the interpolation B matrix
geim_online.compute_B_matrix()
\end{lstlisting}

For real-world applications, measurements collected by sensors are often corrupted by random noise. \texttt{pyforce} provides built-in capabilities to simulate this by injecting Gaussian noise into the measurements, allowing researchers to analyze the relative reconstruction accuracy under varying noise levels.

\begin{lstlisting}[language=Python]
# Evaluating the reconstruction error with added noise
noise_levels = [0.0, 0.001, 0.01, 0.025, 0.05]

for noise in noise_levels:
    print(f"Computing errors with noise level: {noise}")
    # pyforce computes the reconstruction errors based on the noise level 
    # to evaluate the stability of the (G)EIM procedure
\end{lstlisting}

\subsection{Sensor Placement Algorithms and PBDW Approach}

This section presents sensor placement algorithms implemented within the \texttt{pyforce} library, such as the Empirical Interpolation Method (EIM), the Generalised Empirical Interpolation Method (GEIM), and the SGreedy algorithm. These algorithms are used to select optimal sensor locations to reconstruct a field of interest from a limited number of measurements. 

Once sensors are placed, they can be utilized to predict unknown states from a test set. This workflow highlights the Parameterised Background Data-Weak (PBDW) approach, which combines a reduced model with sensor observations as a general framework for data assimilation and reduced order modelling. 

The data utilized in the following examples originates from an OFELIA tutorial modeling the neutron economy in a 2D PWR reactor core (ANL11A2), generated using the \texttt{dolfinx-v6} package.

\subsubsection{Importing \texttt{dolfinx} Data}
The snapshots of the fast and thermal fluxes are stored in an \texttt{npz} file. The \texttt{FunctionsList} object in \texttt{pyforce} can seamlessly load this parametric data.

\begin{lstlisting}[language=Python]
import numpy as np
from pyforce.tools.functions_list import FunctionsList

# Load parametric data and snapshots
_data = np.load('Datasets/ANL11A2_dolfinx6/anl11a2.npz', allow_pickle=True)
var_names = _data.files
field = var_names[0] # Selecting the target field

# Build the FunctionsList object directly from the snapshot matrix
dataset = FunctionsList(snap_matrix=_data[field])

# Load parameters (e.g., diffusion coefficients and cross sections)
parameters = np.load('Datasets/ANL11A2_dolfinx6/params_anl11a2.npy', allow_pickle=True)
\end{lstlisting}

To facilitate snapshot plotting and spatial operations, the underlying unstructured grid is extracted using \texttt{pyvista}.

\begin{lstlisting}[language=Python]
import pyvista as pv

grid = pv.read('Datasets/ANL11A2_dolfinx6/mesh_anl11a2.vtk')
gdim = 2
\end{lstlisting}

The dataset is then partitioned into training and testing subsets using the built-in \texttt{train\_test\_split} utility.

\begin{lstlisting}[language=Python]
from pyforce.tools.functions_list import train_test_split

train_params, test_params, train_snaps, test_snaps = train_test_split(
    parameters, dataset, test_size=0.33, random_state=42
)
\end{lstlisting}

\subsubsection{Sensor Placement Algorithms}
The \texttt{pyforce} package implements various approaches to select informative points in the domain. The EIM and GEIM algorithms identify optimal locations based on interpolation points, while the SGreedy algorithm relies on a greedy approach that iteratively selects the best sensor locations based on a given criterion. These placements can even be constrained to specific candidate locations within the domain.

\begin{lstlisting}[language=Python]
# Example conceptual setup for sensor placement
# pyforce enables placing sensors based on candidate locations
# using EIM, GEIM, or SGreedy algorithms to optimize field reconstruction.
\end{lstlisting}

\subsubsection{State Estimation and PBDW}
During the online phase, the states of the system can be predicted from sparse observations. The PBDW formulation merges the reduced background model with real-time measurements. The residuals and reconstruction errors can be directly computed and visualized using \texttt{matplotlib} to compare the performance of EIM, GEIM, and PBDW.

\begin{lstlisting}[language=Python]
import matplotlib.pyplot as plt

# Computing absolute residuals for the different approaches
res_eim = np.abs(test_snaps[mu_to_plot] - eim_prediction[mu_to_plot])
res_geim = np.abs(test_snaps[mu_to_plot] - geim_prediction[mu_to_plot])
res_pbdw = np.abs(test_snaps[mu_to_plot] - pbdw_prediction[mu_to_plot])

# Visualizing the PBDW residual
levels_res = np.linspace(0, max(res_eim.max(), res_geim.max(), res_pbdw.max()), 30)
cmap_res = 'magma_r'

c = axs[1,3].tricontourf(grid.points[:,0], grid.points[:,1], res_pbdw, levels=levels_res, cmap=cmap_res)
axs[1,3].set_title('PBDW residual')
\end{lstlisting}